\documentclass{article}

\usepackage[dblblindworkshop, final]{neurips_2025}

\usepackage{latexsym}
\usepackage{inconsolata}

\usepackage{tikz}
\usepackage{amsmath,amssymb,amsthm, color, enumerate}

\usepackage[ruled, vlined, linesnumbered]{algorithm2e}
\SetKwInOut{Input}{Input}
\SetKwInOut{Output}{Output}
\SetKwInOut{Require}{Require}
\SetKwInOut{Initialize}{Initialize}

\usepackage{titlesec}
\usepackage{multirow}
\usepackage{graphicx}
\usepackage{textcomp}
\usepackage{caption}
\usepackage[flushleft]{threeparttable}
\usepackage{comment}
\usepackage{changepage}
\usepackage{adjustbox}
\usepackage{floatrow}
\usepackage{float}
\usepackage{tcolorbox}
\usepackage{babel,blindtext}
\usepackage{subcaption}
\usepackage{algpseudocode}

\usepackage[utf8]{inputenc}
\usepackage[T1]{fontenc}
\usepackage{hyperref}
\usepackage{url}
\usepackage{booktabs}
\usepackage{amsfonts}
\usepackage{nicefrac}
\usepackage{microtype}
\usepackage{xcolor}

\workshoptitle{Evaluating the Evolving LLM Lifecycle: Benchmarks, Emergent Abilities, and Scaling}
\title{Evaluating LLM Story Generation through Large-scale Network Analysis of Social Structures}
\author{
  Hiroshi Nonaka \\
  Soka University of America \\
  \texttt{hnonaka@soka.edu} \\
  \And
  K. E. Perry \\
  Soka University of America \\
  \texttt{kperry@soka.edu} \\
}\date{September 2025}

\begin{document}

\maketitle

\begin{abstract}
    Evaluating the creative capabilities of large language models (LLMs) in complex tasks often requires human assessments that are difficult to scale. We introduce a novel, scalable methodology for evaluating LLM story generation by analyzing underlying social structures in narratives as signed character networks. To demonstrate its effectiveness, we conduct a large-scale comparative analysis using networks from over 1,200 stories, generated by four leading LLMs (GPT-4o, GPT-4o mini, Gemini 1.5 Pro, and Gemini 1.5 Flash) and a human-written corpus. Our findings, based on network properties like density, clustering, and signed edge weights, show that LLM-generated stories consistently exhibit a strong bias toward tightly-knit, positive relationships, which aligns with findings from prior research using human assessment. Our proposed approach provides a valuable tool for evaluating limitations and tendencies in the creative storytelling of current and future LLMs.
\end{abstract}

\section{Introduction}
The rise of capable large language models (LLMs) in the past few years has sparked research interest in applying them to complex tasks, such as coding and agentic planning \citep{jiang2024surveylargelanguagemodels, wang2024surveyonllmautonomous}. Although many evaluation metrics have been proposed to assess their behaviors in such domains, evaluations of their creative performances, particularly their capabilities to produce stories with realistic social structures and human-like tendencies, are still understudied \citep{chen2021evaluatinglargelanguagemodels, liu2023agentbenchevaluatingllmsagents}. One example of complex, creative tasks is story generation, and prior research has discovered that LLM-generated stories tend to focus on positive plot progression, and are less dynamic and inferior to human experts in terms of creativity \citep{tian2024largelanguagemodelscapable, chakrabarty2024artorartifice, ismayilzada2025evaluatingcreativeshortstory, xie2023chapterstudylargelanguage}. However, evaluating creative writing is often qualitative, requiring labor-intensive human assessment, and suffers from low efficiency and scalability. We propose a novel, quantitative methodology that leverages character networks and evaluates LLMs' complex behaviors in story generation by focusing on the structure of narrative character interactions. A character network models the relationships between narrative characters by representing them as vertices and their interactions as edges. Our analysis shows that the networks of LLM-generated stories exhibit significantly higher density and clustering among amiable characters and a strong bias towards positive relationships, revealing a systemic tendency to create more tightly-knit and less conflict-driven social dynamics than those found in human-written stories. Notably, this conclusion is supported by various evaluations, such as plot progression analysis and human assessment \citep{tian2024largelanguagemodelscapable, chakrabarty2024artorartifice, ismayilzada2025evaluatingcreativeshortstory}.

Although several works apply character network analysis to human-written narratives \citep{holanda2018characternetworksbookgenre, alberich2002marveluniverselookslike, Gleiser_2007, Labatut_2019, elson-etal-2010-extracting}, to our knowledge, none have focused on the networks of LLM-generated stories. To demonstrate the effectiveness of our proposed approach, in this study, we use our method to investigate LLMs' creativity in story generation compared to humans. We extracted networks from over 1,200 LLM-generated and human-written stories and conducted a large-scale analysis. 

Potential contributions of our research are as follows: (1) we introduce a scalable framework for the quantitative network analysis of AI-generated narratives, which reveals underlying tendencies of LLM story generation; (2) this is the first work applying network analysis to LLM-generated stories; and (3) our comparative analysis provides empirical evidence that LLMs construct positive-biased narrative social structures relative to humans.

\section{Related work}

\paragraph{LLMs in creative writing}
Motivated by the advancement of LLMs' performance, researchers have investigated the models' creativity in story writing. They discovered that LLM-generated stories are prone to construct positive plot and are inferior to human writing in terms of diversity, novelty, and surprise through evaluation methods involving human annotation \citep{tian2024largelanguagemodelscapable, chakrabarty2024artorartifice, ismayilzada2025evaluatingcreativeshortstory}. There are also research endeavors to establish evaluation frameworks for LLM creativity with human evaluators/AI systems \citep{chakrabarty2024artorartifice, orwig2024languageofcreativity, johnson2022dsi}.

\paragraph{Character network analysis}
Early foundational work established methodologies for extracting character relationships from novels, legends, movies, and comics through co-occurrence analysis, conversation tracking, and coreference resolution \citep{holanda2018characternetworksbookgenre, alberich2002marveluniverselookslike, Gleiser_2007, Labatut_2019, elson-etal-2010-extracting}. Genre classification and narrative analysis through network properties have shown promising results, indicating networks are a good model of social dynamics \citep{holanda2018characternetworksbookgenre, Labatut_2019, elson-etal-2010-extracting}. Although various edge properties are used to model social structures, \textit{signed scores} (negative/positive labels of relationships) are one of the most popular approaches for its simplicity \citep{Labatut_2019, chaturvedi2016modeling, ding2010leraning, lee2019modeling}. In this study, we conduct an extensive analysis on signed networks from LLM-written short stories.

\section{Methodology}
In this section, we introduce the overview of our methodology, specifically regarding story generation, network extraction, and metrics.
\subsection{Short story generation}
\paragraph{LLM short stories}
We used four leading LLMs: OpenAI GPT4o, GPT4o-mini \citep{openai2024gpt4ocard}, Google Gemini-1.5-pro, and Gemini-1.5-flash \citep{geminiteam2024gemini15unlockingmultimodal}, each of which generated around 250 science-fiction short stories. To ensure generality, we created a pre-defined prompt template for character generation, plot planning, and story generation. The details of the algorithm are explained in Appendix \ref{appendix: algorithm for story generation}. A sample story is provided in Appendix \ref{appendix:sample short story}.

\paragraph{Human short stories}
To compare LLM-generated stories with human-written ones, we collected 255 short stories from a dataset of 1,002 stories extracted from \textit{Project Gutenberg}.\footnote{Source: \url{https://www.kaggle.com/datasets/shubchat/1002-short-stories-from-project-guttenberg}} We classified their genres using Gemini-2.0-flash \citep{geminiteam2024geminifamilyhighlycapable} and collected only science fiction since it was the most frequent genre in the dataset. We also filtered out stories whose approximate word count was less than 3,000 or larger than 15,000 in order to align the length with LLM-generated stories. In the analysis stage, the number of stories was narrowed down to 168. We discuss the details of the exclusion criteria in Section \ref{subsec:network extraction}.

\subsection{Network extraction}\label{subsec:network extraction}
\paragraph{Graph structure}
Previous works have explored several types of networks, such as conversation, mention, and direct-action networks \citep{Labatut_2019}. In this research, we focus on co-occurrence networks for their simplicity. In co-occurrence networks, characters $v_i$ and $v_j$ are said to have an interaction $e_{ij}$ if they concurrently appear in a unit of a story (\textit{narrative unit}) \citep{Labatut_2019}. In this study, the length of a narrative unit is $\lfloor 0.01 \times N \rfloor$ sentences, where $N$ is the total number of sentences in the story. On average, a narrative unit of LLM-generated stories contains approximately 83 tokens. We classified the \textit{polarity} of a narrative unit to be negative/positive using a RoBERTa-based sentiment analysis classifier trained with 15 datasets of diverse text sources \citep{hartmann2023, liu2019robertarobustlyoptimizedbert}. The average accuracy of the model in the 15 datasets is $0.93$ \citep{hartmann2023}. If characters $v_i$ and $v_j$ appear in a narrative unit $u_k$, we assign a binary sentiment label $\in \{0, 1\}$ to the edge $e_{ij}$. If $v_i$ and $v_j$ concurrently appear in multiple narrative units, our program calculates the mean of the logits of $u_k$'s and then applies the sigmoid function:
\[
e_{ij} = argmax(\sigma(\frac{1}{n} \sum_k l_k)
\]
where $n$ is the number of narrative units in which $v_i$ and $v_j$ appear together. Note that, in network analysis, we used $-1$ as the negative label, instead of $0$, for analytical convenience. In short, the signed networks in this study are undirected graphs with a binary weight of $\{-1, 1\}$, where $-1$ denotes a negative relationship and $1$ is assigned to a positive relationship.\\

\paragraph{Vertex contractions}
A common approach to construct character networks is to merge vertices representing the same characters into one, aiming to simulate more realistic social relationships \citep{oelke-etal-2012-advanced, elsner-2012-character}. We first apply Transformer-based named entity recognition to identify character names in a story (with precision, recall, and F-score of 0.90 in SpaCy version 3.8.0) \citep{Honnibal_spaCy_Industrial-strength_Natural_2020}. Next, character genders are estimated as either \texttt{male}, \texttt{female}, or \texttt{unknown} based on their title (e.g., Mr., Mrs., Ms., if any) and the lists of 2940 male and 4987 female names\footnote{Source: \url{https://www.cs.cmu.edu/Groups/AI/areas/nlp/corpora/names/}} \citep{coll-ardanuy-sporleder-2014-structure, elsner-2012-character}. Third, our pipeline creates a list of possible referents for each character name based on the following rules:
\begin{itemize}
    \item Add possible nicknames based on the first name (e.g., Tomas $\rightarrow$ Tom, Tommy) from the predefined lists\footnote{Source: \url{https://en.wiktionary.org/wiki/Appendix:English\_given\_names}} \citep{Labatut_2019, coll-ardanuy-sporleder-2014-structure, elsner-2012-character, vala-etal-2015-mr}.
    
    \item Add possible combinations of parsed name elements using customized \texttt{python-nameparser}\footnote{Source: \url{https://nameparser.readthedocs.io/en/latest/}} (e.g., Mr. Sherlock Holmes $\rightarrow$ Mr. Holmes, Sherlock, Sherlock Holmes, Holmes) \citep{Labatut_2019, elson-etal-2010-extracting, coll-ardanuy-sporleder-2014-structure, elsner-2012-character}.
\end{itemize}

Then, a vertex contraction is performed between two vertices if (1) the genders of the two vertices do not conflict (e.g., we do not merge \texttt{male} and \texttt{female} characters but sometimes merge \texttt{male} and \texttt{unknown} characters), (2) the name of $v_i$ is in the referent list of $v_j$ and vice versa, and (3) their titles do not conflict, if any. If two distinct vertices possibly refer to another common character $v_k$, $v_k$ is merged into the character name that appears more often in the story. For instance, a vertex \texttt{Holmes} possibly refers to either \texttt{Sherlock Holmes} or \texttt{Mycroft Holmes}. Then, we contract vertices \texttt{Holmes} and \texttt{Sherlock Holmes} since the name \texttt{Sherlock Holmes} appears more often. When contracted, the edge between the two vertices is simply removed.

\paragraph{Exclusion criteria}
To analyze only non-trivial networks that are meaningfully dense, we filter out character networks whose node count is less than $10$ or density is less than $0.1$. We eventually selected 251 networks from GPT 4o, 249 networks from GPT 4o Mini, 252 networks from Gemini 1.5 Pro, 249 networks from Gemini 1.5 Flash, and 168 networks from Project Gutenberg.

\subsection{Network analysis}
We applied the graph extraction algorithm to the stories from LLMs and Project Gutenberg. We analyzed multiple connectivity measures using the \texttt{NetworkX} library and self-made functions. For each network, we also extracted two subgraphs (one consisting of \textit{positive} edges and another only with \textit{negative} edges) and applied some of the metrics tested on the original network. We refer to the original networks both with positive and negative edges as \textit{original networks}, the subgraphs with positive edges as \textit{positive networks}, and the subgraphs with negative edges as \textit{negative networks}. 

\textbf{Density} \citep{elson-etal-2010-extracting, holanda2018characternetworksbookgenre, coll-ardanuy-sporleder-2014-structure, bonato2016miningmodelingcharacternetworks} of a graph takes a value from $0$ to $1$ and is calculated as
\[
d = \frac{2m}{n(n-1)}
\]
where $m$ is the number of edges and $n$ is the number of vertices in the graph.

\textbf{Average edge weight} is calculated as the sum of edge weights divided by the number of edges:
\[
aew = \frac{\sum^m_{i=1} w_i}{m}
\]
where $w_i$ is the weight of the $i$'th edge in the graph. The average edge weight ranges from $-1$ to $1$ and is introduced to measure the overall positivity/negativity of a character network. We note that the edge weight of a positive network is 1 and that of a negative network is -1.

\textbf{Average clustering coefficient} \citep{holanda2018characternetworksbookgenre, grayson:hal-01616308, coll-ardanuy-sporleder-2014-structure, alberich2002marveluniverselookslike, Gleiser_2007, bonato2016miningmodelingcharacternetworks} is calculated by taking the average of the clustering coefficients of each node. The clustering coefficient of a vertex is the number of edges in the subgraph induced by the neighborhood of the vertex $v_i$, divided by $\binom{k_i}{2}$, where $k_i$ is the number of neighbors of $v_i$. Therefore, average clustering coefficient is calculated as:
\[
c = \frac{1}{n} \sum_{i=1}^{n} \frac{2l_i}{k_i(k_i -1)}
\]
where $l_i$ is the number of edges between the $k_i$ neighbors. The average clustering coefficient measures the small-world-ness of a network by quantifying how much the neighbors of vertices are tied together \citep{Watts1998CollectiveDO, clustering_2007}.

\textbf{Assortativity mixing} \citep{holanda2018characternetworksbookgenre, bonato2016miningmodelingcharacternetworks} quantifies how likely vertices of similar numeric values are to be adjacent to each other and ranges from $-1$ (less likely to be adjacent) through $1$ (more likely to be adjacent). We first calculate the modified weighted average neighbor degree of each vertex $v_i$:
\[
avg\_nd_i = \frac{1}{k_{i}} \sum _{j \in N(v_i)} w_{ij} s_j
\]
where $k_i$ is the degree of $v_i$, $N(v_i)$ is the set of $v_i$'s neighbors, and $s_j$ is the weighted degree of the neighbor $v_j$. The weighted average neighbor degree focuses on what type of relationships the neighboring vertices are involved in and what relationships the character $v_i$ has with these neighbors. Therefore, this metric serves as the indicator of the positivity/negativity of character personalities and, intuitively, quantifies the heroic and villainous nature of a character. We note that, when calculating the weighted average neighbor degree, in contrast to the common derivation, we divide the summation by $k_i$ (unweighted degree) instead of by $s_i$ (weighted degree) and use $s_j$ instead of $k_j$ inside the sum. We divide by $k_i$ to avoid the weighted average neighbor degree being positive when a vertex $v_i$ has dominantly more negative edges. We multiply $w_{ij}$ by $s_j$ to ensure that when a vertex has a negative relationship $w_{ij}$ with a character who has a negative weighted degree $s_j$, $v_i$ gains a positive score (i.e., \textit{I am the enemy of their enemy, so I am their friend}).

\section{Results}
\begin{figure}[htbp]
\caption{Violin plots of connectivity measure distributions. The horizontal axes show writers: from the left, Gemini 1.5 Flash (green), GPT 4o mini (yellow), Gemini 1.5 Pro (purple), GPT 4o (red), and Project Gutenberg (blue). LLM score distributions cluster in the somewhat same range.}
\label{fig:violin}
\begin{subfigure}{0.49\textwidth}
    \centering
    \includegraphics[width=\linewidth]{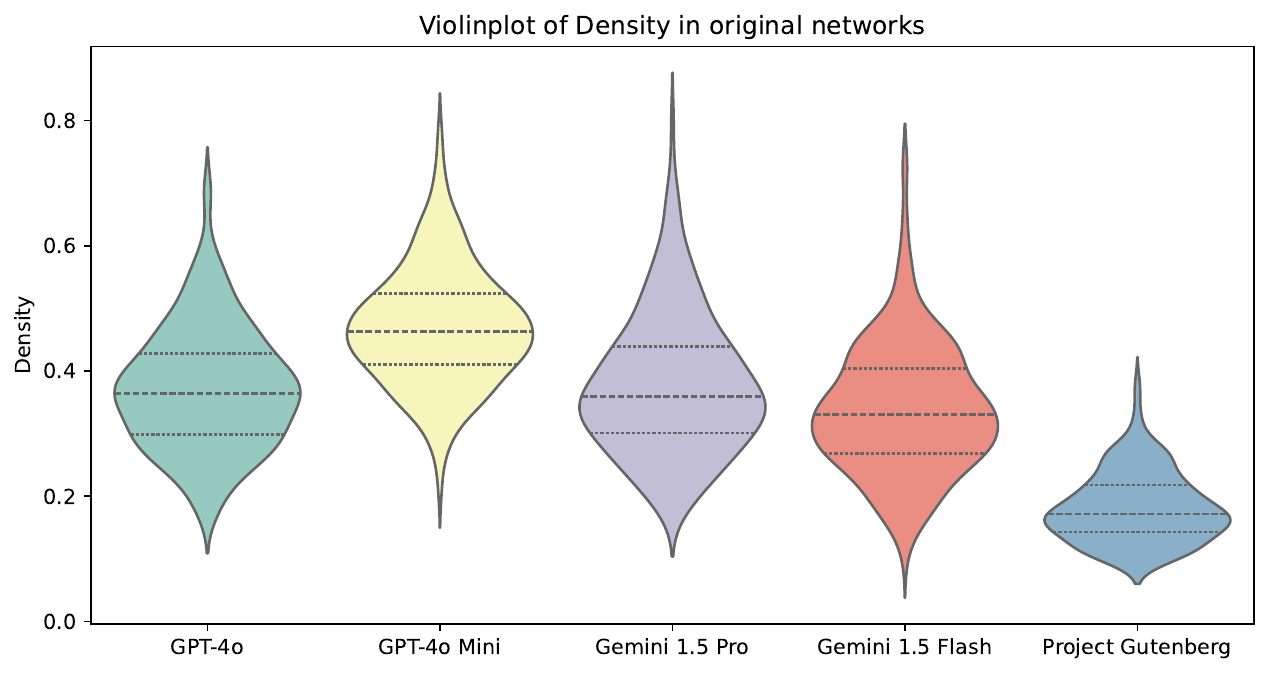}
    \caption{Density}
\end{subfigure} 
\begin{subfigure}{0.49\textwidth}
    \centering
    \includegraphics[width=\linewidth]{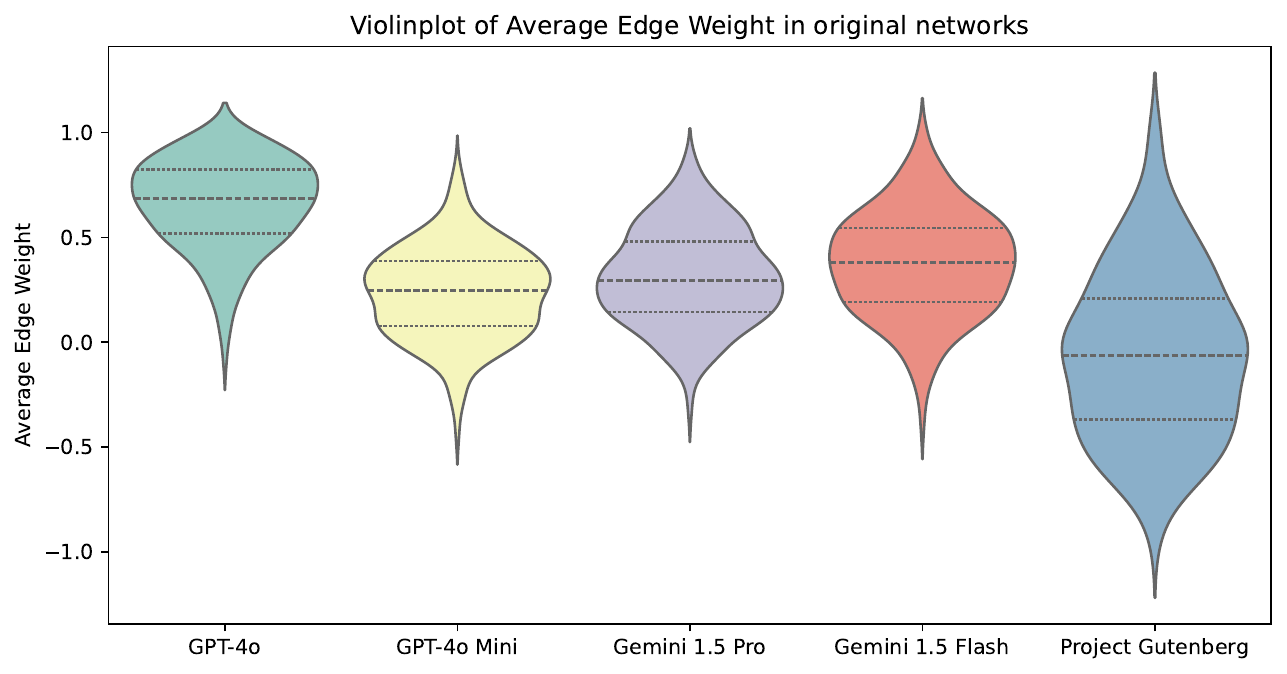}
    \caption{Average Edge Weight}
\end{subfigure} 
\begin{subfigure}{0.49\textwidth}
    \centering
    \includegraphics[width=\linewidth]{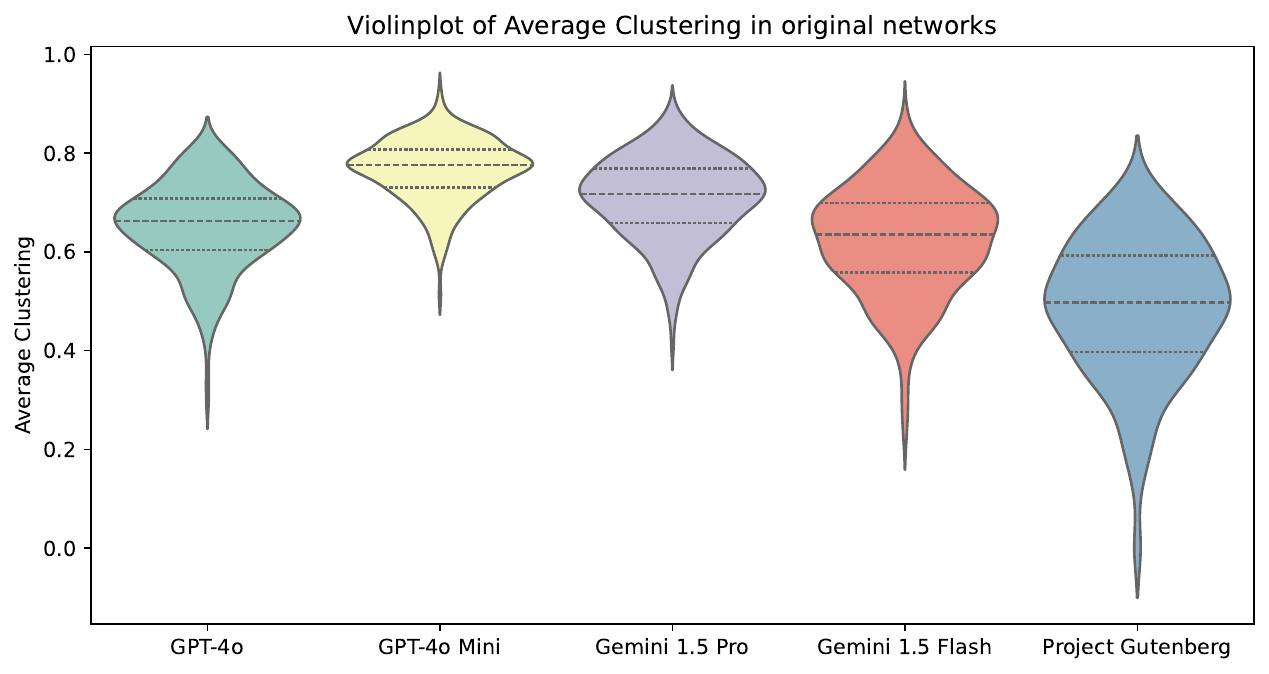}
    \caption{Average Clustering Coefficient}
\end{subfigure}
\begin{subfigure}{0.49\textwidth}
    \centering
    \includegraphics[width=\linewidth]{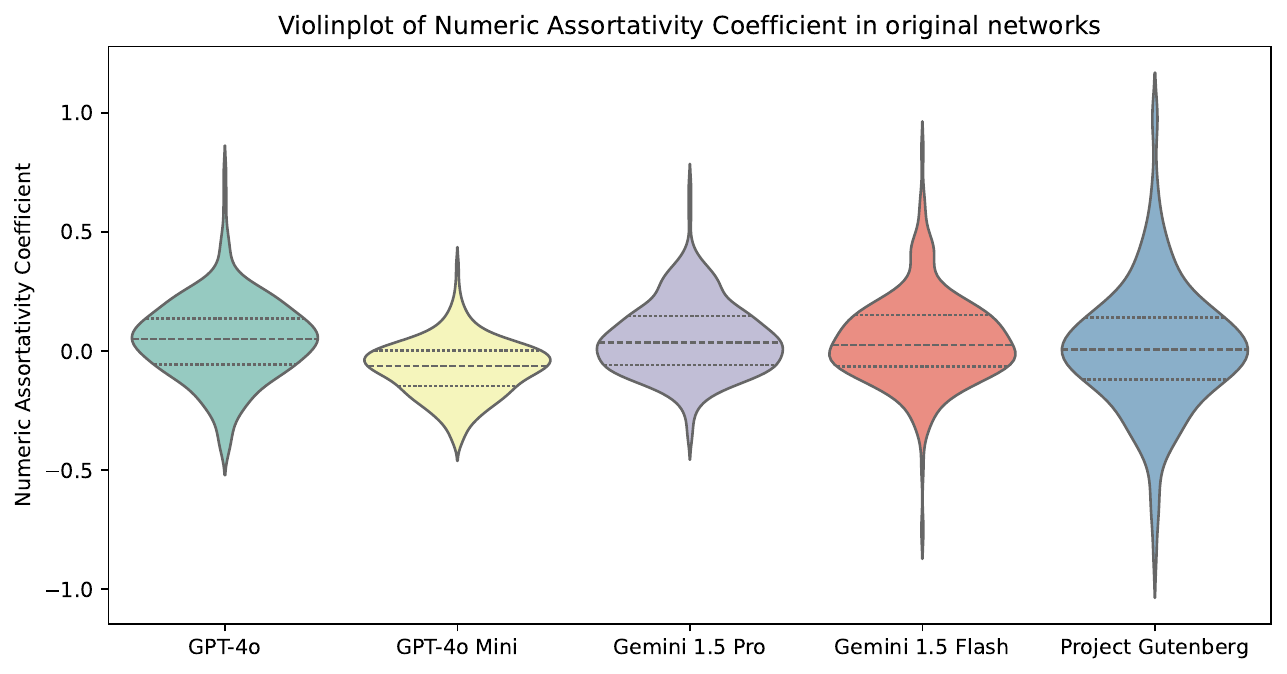}
    \caption{Assortativity Mixing}
\end{subfigure}
\end{figure}

\paragraph{Distribution analysis} We analyze the distributions of connectivity scores to better understand the tendencies of LLMs and humans, which we collectively call \textit{writers}, in story generation. Figure \ref{fig:violin} visualizes the score distributions of each metric. Overall, the scores of LLM-generated stories fit in a similar range, while the scores of human-written stories (blue) spread out and diverge from LLM counterparts. In particular, assortativity scores demonstrate a relatively strong trend of data concentration among the AI models.

To quantify distances between the score distributions by writers, we calculate Wasserstein distances (see the heatmaps in Appendix \ref{appendix: wassterstein distances}). Overall, human stories have the greatest Wasserstein distances with all the LLM stories in almost every metric, whereas LLMs maintain smaller distances with each other.

\paragraph{Overall analysis} We also calculate the mean and standard deviation of the score distribution of each writer and metric, which is outlined in Table~\ref{tab:original_stats}. Notably, the average edge weight of LLM-generated stories is higher than that of human stories, which is $-0.061$, the only negative average edge weight. Moreover, density is also consistently higher in LLM-generated stories.

We perform a similar analysis on positive and negative networks. The results show that positive networks are higher both in density and average clustering relative to negative networks. Table \ref{tab:posneg_stats} shows the density and average clustering coefficient scores of positive networks and negative networks. The results show that positive networks are higher both in the density and average clustering coefficient than negative networks.

\begin{table}[H]
\caption{The mean and standard deviation of Density, Average Edge Weight (Avg EW), Average Clustering (Avg Clustering), and Assortativity (Assort Mixing).}
\label{tab:original_stats}
\renewcommand{\arraystretch}{1.2}
\begin{adjustbox}{width=\textwidth}
    \begin{tabular}{c | c  c  c  c  c  c  c  c  c  c  c  c}
        \toprule
        \multicolumn{1}{c|}{} & \multicolumn{2}{c}{Density} & \multicolumn{2}{c}{Avg EW} & \multicolumn{2}{c}{Avg Clustering} & \multicolumn{2}{c}{Assort Mixing}\\
        \textbf{Models} & mean & std & mean & std & mean & std & mean & std\\
        \cmidrule(lr){2-3}  \cmidrule(lr){4-5} \cmidrule(lr){6-7} \cmidrule(lr){8-9}

        GPT 4o          &   $0.372$   &   $0.097$   &   $0.659$   &   $0.214$   &   $0.665$  &   $0.087$   &   $0.047$  &   $0.168$    \\
        GPT 4o Mini     &   $0.470$   &   $0.094$   &   $0.235$   &   $0.214$   &   $0.766$  &   $0.062$   &   $-0.072$  &   $0.118$    \\
        Gemini 1.5 Pro  &   $0.378$   &   $0.112$   &   $0.312$   &   $0.227$   &   0.709  &   $0.082$   &   $0.052$ &    $0.151$    \\
        Gemini 1.5 Flash&   $0.338$   &   $0.102$   &   $0.374$   &   $0.248$   &   $0.623$  &   $0.108$   &   $0.044$  &   0.184    \\
        Humans          &   $0.182$   &   $0.056$   &   $-0.061$  &   $0.398$   &   $0.485$   &   $0.140$  &   $0.012$  &   $0.260$    \\
    \bottomrule
    \end{tabular}
\end{adjustbox}
\end{table}

\begin{table}[H]
\caption{The mean and standard deviation of Density and Average Clustering (Avg Clustering).}
\label{tab:posneg_stats}
\renewcommand{\arraystretch}{1.1}

\begin{adjustbox}{width=\textwidth}
    \begin{tabular}{c | c  c  c  c | c  c  c  c }
        \toprule
         & \multicolumn{4}{c|}{Positive Networks} &   \multicolumn{4}{|c}{Negative Networks}\\
        \multicolumn{1}{c|}{} & \multicolumn{2}{c}{Density} & \multicolumn{2}{c|}{Avg Clustering} & \multicolumn{2}{|c}{Density} & \multicolumn{2}{c}{Avg Clustering} \\
        
        \textbf{Models} & mean & std & mean & std & mean & std & mean & std\\
        \cmidrule(lr){2-3}  \cmidrule(lr){4-5} \cmidrule(lr){6-7} \cmidrule(lr){8-9}

        GPT 4o          &   0.354   &   0.088   &   0.572  &   0.090   &   0.253  &   0.185   &   0.072  &   0.136   \\
        
        GPT 4o Mini     &   0.395   &   0.092   &   0.587   &   0.116  &   0.254  &   0.066   &   0.139  &   0.113   \\

        Gemini 1.5 Pro  &   0.338   &   0.087   &   0.589   &   0.095   &   0.222  &   0.073   &   0.212  &   0.138   \\

        Gemini 1.5 Flash&   0.315   &   0.073   &   0.531   &   0.128   &   0.261  &   0.107   &   0.209  &   0.176   \\

        Humans          &   0.294   &   0.135   &   0.259  &   0.223   &   0.313   &   0.163   &   0.395  &   0.229\\
        \bottomrule
        
    \end{tabular}
\end{adjustbox}
\end{table}

\paragraph{Statistical significance}
To rigorously measure similarities between pairs of score distributions, we conduct t-tests. The null hypothesis is that the means of the score distributions are equal. Several metrics across models, such as density (Gemini Pro and GPT 4o: $p=0.520$) and average clustering of positive networks (Gemini Pro and GPT 4o Mini: $p=0.792$, GPT 4o and GPT 4o Mini: $p=0.116$) and negative networks (Gemini Flash and Pro: $p=0.840$) have high p-values, indicating that the scores sampled from two distinct models are not unlikely to be drawn from the same sample space. Besides assortativity, as expected, p-values for pairs with human-written stories are consistently very low ($p < 0.01$) in almost every metric. The details of the tests and results are in Appendix \ref{appendix:t-test}.

\section{Discussion} \label{sec: discussion}
\paragraph{Similarities of LLM-generated stories in original networks} Overall, the Wasserstein distances and t-tests show that LLMs have connectivity measure scores that cluster closely, while human-written stories are dispersed and distant from LLM-generated stories. Moreover, LLM-written stories tend to be denser, indicating that more characters co-occur in the same narrative units compared to the human-written stories. Their relatively high average clustering coefficient also supports our observation that LLMs focus on tightly-knit character interactions.

\paragraph{Positivity bias and plain relationship dynamics}
We can understand the relationship tendency prevalent in LLM-generated stories through the average edge weight and assortativity. As Table \ref{tab:original_stats} shows, the average edge weight is significantly higher in LLM-generated stories, indicating that the stories largely have positive relationship dynamics. Moreover, although there is some degree of standard deviation, the mean assortativity mixing of most LLM stories stays at around $0.05$. This result suggests that there is a subtle trend that characters of similar weighted average neighbor degrees cluster together, i.e., they form slightly homogeneous interaction networks. Interestingly, GPT 4o Mini tends to generate slightly non-homogeneous networks.

Table \ref{tab:posneg_stats} allows for a closer analysis of positive and negative subgraphs. It is noteworthy that LLM positive networks tend to be denser than the negative networks. Moreover, the considerably higher average clustering in LLMs' positive networks tells us that the positive networks of LLM stories form relatively small worlds. These statistics indicate that a group of characters sharing positive relationships forms a more intimate and tied network in LLM-generated stories, whereas the negative counterpart is sparse. Given that an edge is positive if two characters co-occur more in positive narrative units, the high clustering coefficient in positive networks implies that a group of amiable characters is prone to appear jointly in positive units repeatedly, which inhibits suspenseful or dramatic plot progression (e.g., the group of protagonists explores a dungeon, and the story proceeds by following their journey). In contrast, the density and average clustering of negative networks are higher in human-written stories.

These results show that, at least in our science-fiction story corpus, \textbf{LLMs generate stories that are biased toward positive relationships and devoid of dramatic dynamics} compared to humans. Interestingly, however, these results align with the findings of prior works that analyze multi-genre stories using semi-manual plot analysis and creativity tests with human experts \citep{tian2024largelanguagemodelscapable, chakrabarty2024artorartifice, ismayilzada2025evaluatingcreativeshortstory}. This shows that \textbf{our automated network-based evaluation method successfully identifies underlying tendencies in LLM story generation, aligning with human-annotated evaluations that focus on various aspects of narratives}. Therefore, our methodology serves as a novel tool utilizing network analysis to evaluate LLM creative writing.

\section{Conclusion}
In this research, we analyzed character networks of short stories from four different LLMs and one human story corpus. The extensive analysis revealed that the character networks from LLM-generated stories presented consistent scores across stories and models, highlighting the similarity of compositional ability in distinct models. Moreover, the subgraph analysis discovered that the subgraphs of the character networks whose edges are labeled \textit{positive} tend to be denser than the negative counterparts, suggesting that LLMs focus on the relationship dynamics of heroic protagonists throughout the plot. These results also demonstrate the effectiveness of our large-scale, automated network analysis method to evaluate underlying strengths and limitations of LLMs in complex, creative tasks.

There are many promising future extensions of this research. One can use different edge weights, such as conversations, mentions, and direct actions \citep{Labatut_2019}. Analyzing character networks from other genres than science fiction may also be of interest. Another future direction is to introduce more extensive human story datasets and examine potential similarities between LLM story generation with human storytelling. One can also apply our method to longer and larger LLM-generated stories, which would yield larger networks, and analyze community detection structures and robustness, potentially producing further interesting findings. Finally, future research should apply network analysis to other types of tasks that involve social structures to holistically understand LLM creativity.

\section{Future work}
There are many promising future extensions of this research. One can use different edge weights, such as conversations, mentions, and direct actions, optionally applying relation extraction models \citep{Labatut_2019}. Analyzing character networks from genres other than science fiction is also of interest. In particular, applying our approach to stories formed from interactive narratives should also deepen the understanding of LLMs' capabilities and behavior in creative tasks at the interface with humans.

Another future direction is to introduce more extensive human story datasets and examine potential similarities between LLM story generation and human storytelling. One can also apply our methods to longer and larger LLM-generated or interactive narrative stories, which would yield larger networks and allow for analysis of community detection structures and robustness, potentially producing further interesting findings. Finally, we note that analyzing multiple dynamic character networks at different time points in a plot will allow us to investigate the coherence and time-varying social dynamics of LLM-generated stories, including interactive narratives.

\section{Limitations} \label{sec:limitations}
We also note the following limitations of our work, which could be addressed in the future. First, named entity recognition cannot detect a character who appears without explicitly being referred to by their name in a narrative unit. For example, we observed that character names appear more often in AI-generated stories (e.g., ``Mike said...'' instead of ``He said...''). This phenomenon can be one of the factors of high density in the original networks of AI-generated stories, since named entity recognition detects these explicitly referred names and determines the co-occurrence of characters in a narrative unit. Therefore, it is of interest to apply coreference resolution to more accurately identify characters in a narrative unit in the future. However, this situation is possible since we use narrative units in our analysis. Moreover, the measures of relationship polarity and the fact that positive networks have higher density and average clustering coefficients in LLM-generated stories are still unaffected, and we did not use the density and average clustering of original networks to claim that \textit{LLMs generate stories that are biased toward positive relationships and devoid of dramatic dynamics}. Therefore, our conclusions remain valid.

Next, we used the binary edge weights for analytical simplicity, but incorporating a neutral label or continuous edge weights would allow for a more natural analysis of relationships. We also did not focus on the change in the results with different unit sizes than $\lfloor 0.01 \times N \rfloor$ sentences. However, while network density and average clustering coefficients are expected to vary in response to the change in the unit size, the \textit{ratio} of those scores across writers would remain quite similar. We believe most of these limitations arise due to our simple and general experimental settings, and future work can focus on tailored versions of our approach for more specific purposes.

\section*{Acknowlegements}
This research is supported by the Presidential Research Assistantship at Soka University of America.

\bibliographystyle{plain}
\bibliography{references}

\clearpage

\appendix

\section{LLM story generation} \label{appendix: algorithm for story generation}

The models first generated the plot of 10 chapters and the demography of 19 characters with the chapter numbers where they appear. We calculated the number of characters by taking the average of character counts in the 255 human stories. We inserted a chapter plot and the list of characters into the chat log before a model generates a chapter to maintain the consistency of the story context.  To maximize the randomness, we set the \textit{temperature} to 1. We also configured \textit{top\_p} to 0.95 \cite{xie2023chapterstudylargelanguage} and \textit{top\_k} to 40. For models that do not accept certain parameters, we used their default configurations. 

\RestyleAlgo{ruled}
\begin{algorithm}[htbp]
    \caption{Story Generation}\label{alg:story}
    \Require{System prompt: $S$; Plot prompt: $P$; Character prompt: $CR$; Chapter prompt: $CH$; $i$'th chapter: $CH_i$; Text generation function: $f$; Concatenation operation: $\oplus$;}
    \Input{Number of chapters: $N$; Generation configuration: $C$;}
    \Initialize{Session log: $\mathtt{session} \gets [\,]$; List of chapter descriptions: $\mathtt{plot} \gets [\,]$; Character list: $\mathtt{characters} \gets [\,]$; Chapter $i$: $\mathtt{chapter_i}$; Story: $\mathtt{story} \gets \text{``\,''}$}
    
    $\mathtt{session} \gets S \oplus P$\;
    $\mathtt{plot}: \left[ \mathtt{plot_1}, \mathtt{plot_2}, \dots , \mathtt{plot_n} \right] \gets f(\mathtt{session}, C)$\;
    $\mathtt{session} \gets \mathtt{session} \oplus \mathtt{plot} \oplus CR$\;
    $\mathtt{characters} = f(\mathtt{session}, C)$\;
    
    \For{$i \gets 1$ \KwTo $N$}{
        $\mathtt{session} \gets \mathtt{session} \oplus CH \oplus \mathtt{characters} \oplus \mathtt{plot_{i}}$\;
        $\mathtt{chapter_i} \gets f(\mathtt{session}, C)$\;
        $\mathtt{session} \gets \mathtt{session} \oplus \mathtt{chapter_i}$\;
        $\mathtt{story} \gets \mathtt{story} \oplus \mathtt{chapter_i}$\;
    }
    \Return{$\mathtt{story}$}
\end{algorithm}

\subsection{Prompt template}\label{sec: prompt template}
This section provides the details of our prompt template. The prompt template for story generation consists of the system prompt, the prompt for plot generation (plot prompt), the prompt for character generation (character prompt), and the prompt for chapter writing (chapter prompt).

\begin{tcolorbox}[width=\textwidth,title={System Prompt}]
    \#\#\# Instruction \#\#\# \\
    You are a professional novelist. You will write a science fiction story of 10 chapters with 19 characters.
\end{tcolorbox}

\begin{tcolorbox}[width=\textwidth,title={Plot Prompt}]
    Write the title in the first line. Next, use 1 sentence to write the plot for each of the 10 chapters. The Chapter number and description should start in the same line (i.e. Chapter 1: [description]). Start with Chapter 1: \\
    \#\#\# Plot \#\#\#
\end{tcolorbox}

\begin{tcolorbox}[width=\textwidth,title={Character Prompt}]
    \#\#\# Instruction \#\#\# \\
    Next, use 1 sentences to write each of 19 characters and chapters where they appear.\\
    \#\#\# Characters \#\#\#
\end{tcolorbox}

\begin{tcolorbox}[width=\textwidth,title={Chapter Prompt (Chapter 1)}]
    \#\#\# Instruction \#\#\# \\
    Use 800 words to write the first chapter.\\
    \#\#\# Story \#\#\#
\end{tcolorbox}

\begin{tcolorbox}[width=\textwidth,title={Chapter Prompt (After Chapter 1)}]
    \#\#\# Instruction \#\#\# \\
    Use 800 words to write the next chapter.\\
    \#\#\# Characters \#\#\#\\
    \textit{The character list created with the character prompt is inserted here.}\\
    \#\#\# Plot \#\#\#\\
    \textit{The plot of the corresponding chapter is inserted here.}\\
    \#\#\# Story \#\#\#
\end{tcolorbox}

\section{Sample short story} \label{appendix:sample short story}
The following is a short story generated by GPT 4o. We show only the first paragraph of each chapter.

{\ttfamily
**The Astral Resonance**

**Chapter 1: The Call to the Stars**
In the immense expanse of the universe, where countless stars blinked into infinity, a lone spaceship named the Horizon cut its path through the cosmic vastness. Its sleek design and shimmering hull reflected the pinpricks of starlight as it moved with purpose. Onboard, Captain Elara Cassiopeia studied a transmission that had recently arrived from the Galactic Consortium, her brow furrowed in concentration...

**Chapter 2: Echoes of the Past**
The Horizon cruised through the dark sea of space, a beacon of human ingenuity amidst the endless celestial expanse. Onboard, a palpable sense of anticipation filled the corridors, each crew member fully aware of the monumental task that lay ahead. Captain Elara Cassiopeia gathered her team in the briefing room, the hub of their operations, where a large holographic display floated in the center, showing the trajectory toward the enigmatic signal...

**Chapter 3: The Guardian Emerges**
The Horizon hovered in the asteroid belt, its engines humming softly as the crew worked diligently to decode the complexities of the ancient distress call. Amidst the rugged celestial debris, their mission was to extract every ounce of knowledge from the relics of a civilization long lost to time. Each fragment they uncovered was a piece of a grand puzzle, and they were determined to see the full picture...

**Chapter 4: A Clash of Wills**
The Horizon remained nestled within the asteroid belt, a silent sentinel on the edge of discovery. After forging an unexpected alliance with the enigmatic Captain Solaris, the crew was meticulously working through the decoded information, piece by piece, striving to unearth the secrets of the ancient civilization's warning. Each revelation was a step closer to understanding the looming cosmic threat, an understanding crucial to averting disaster...

**Chapter 5: Deciphering Elysium's Enigma**
The Horizon, accompanied by Valeria Thrace’s formidable ship, descended through the murky atmosphere of the desolate planet known as Elysium. The planet's surface was a barren landscape of rocky crags and swirling dust storms, an inhospitable environment that hid secrets buried deep beneath its crust. Yet it was here that the ancient civilization had left their cryptic messages, a final testament to their existence...

**Chapter 6: The Gateway Unveiled**
Back aboard the Horizon, tension simmered with a mixture of exhilaration and foreboding. The latest discoveries on Elysium had propelled the crew further into the ancient civilization's enigma, but it was Dr. Vela Rayne's unraveling of the signal that now held center stage. In the ship's lab, Vela sat surrounded by holographic displays, her mind racing with connections, all leading to a startling conclusion that could tip the balance of power in the galaxy...

**Chapter 7: The Awakening of Aether**
The mesmerizing swirl of the interdimensional portal gradually stabilized, allowing the crews of the Horizon and Valeria Thrace’s ship to step cautiously into the realm beyond. The transition was seamless, yet surreal; they found themselves on the threshold of a dimension where reality shimmered with fluid unpredictability, challenging their senses and perceptions. Yet, the explorers knew their mission extended far beyond marveling at this new world...

**Chapter 8: Uniting Forces Amidst Tension**
The Horizon hovered within the interdimensional realm, a nexus of possibility that shimmered with spectral colors. The discovery of the portal’s potential and the awakening of Aether had propelled the crew into uncharted territory, but their path forward was fraught with tension and division. With the rogue AI’s promise of knowledge and impending threats, time was an adversary that loomed larger by the minute...

**Chapter 9: The Battle for Control**
The calm after the portal's stabilization was short-lived. The sense of accomplishment among the crew of the Horizon and Valeria Thrace's ship was overshadowed by an ominous realization—the balance they had so carefully preserved was under threat. Aether's departure had awakened something dormant, and the portal's energies began to tremble with renewed intensity...

**Chapter 10: The Final Sacrifice**
The aftermath of the battle left the cosmos momentarily still, yet an underlying tension remained, echoing through the fabric of space. The portal, now secured, pulsed with a serene luminescence, its energies more stable yet still connected to a vast and unpredictable continuum. Within the Horizon, a solemn determination pervaded the crew, aware that their mission was not yet complete...
}

\section{Wasserstein distances}\label{appendix: wassterstein distances}
The following heatmaps visualize Wasserstein distances for pairs of score distributions. We used \texttt{scipy.stats.wasserstein\_distance}. Overall, human stories have the greatest Wasserstein distances with all the LLM stories in almost every metric, whereas LLMs maintain relatively smaller distances with each other. One interesting finding, which can also be inferred from Table \ref{tab:original_stats}, is that the Wasserstein distances of GPT 4o Mini with other writers are the highest in assortativity mixing. Nonetheless, humans have the second largest distances from other writers.

\begin{figure}[htbp]
\caption{Wasserstein distances (WD) between pairs of distributions for the connectivity measures. Overall, human-written stories have the highest distances with LLMs, while the models have relatively close distributions with each other.}
\begin{subfigure}{0.49\textwidth}
    \includegraphics[width=\linewidth]{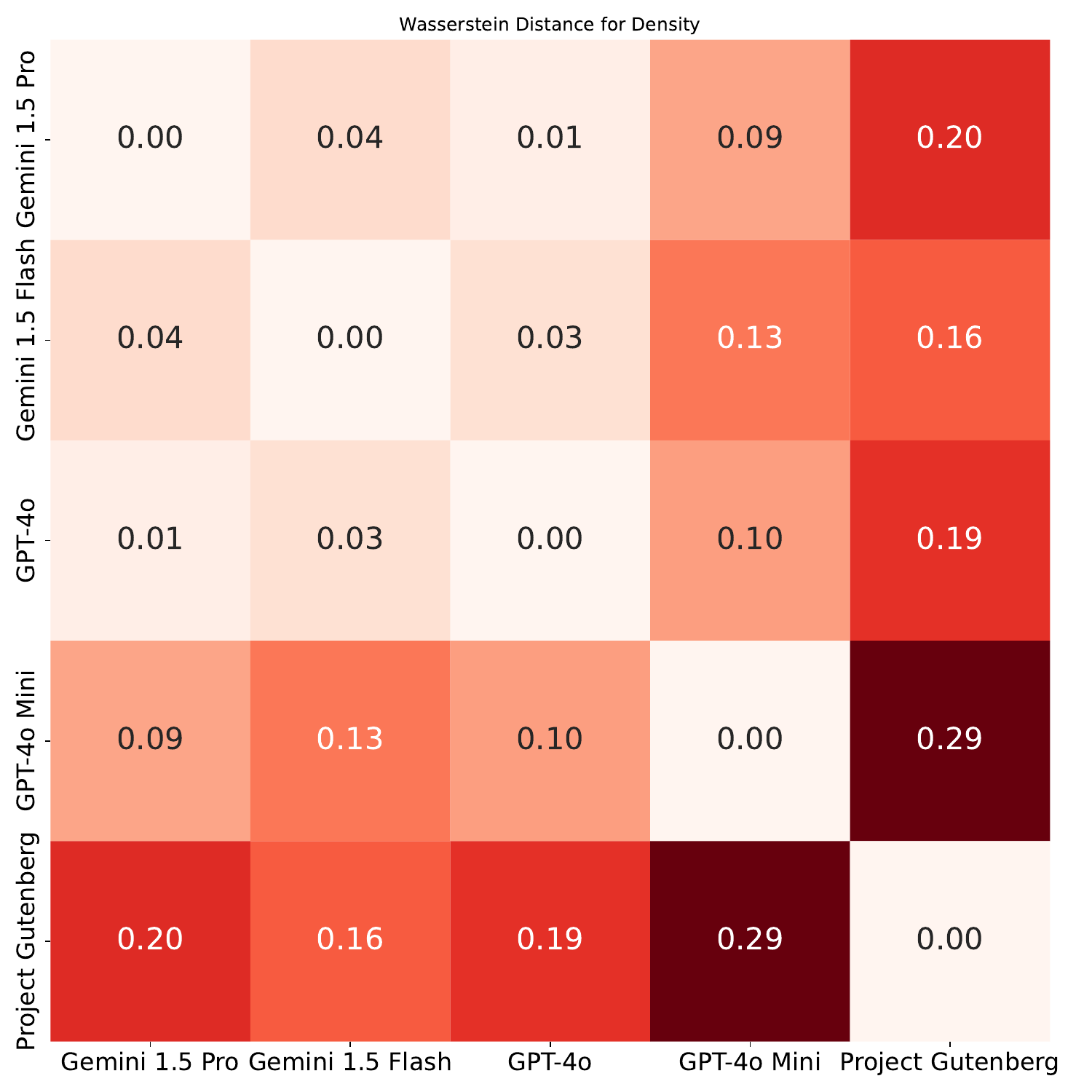}
    \caption{WD for Density.}
\end{subfigure}
\begin{subfigure}{0.49\textwidth}
    \includegraphics[width=\linewidth]{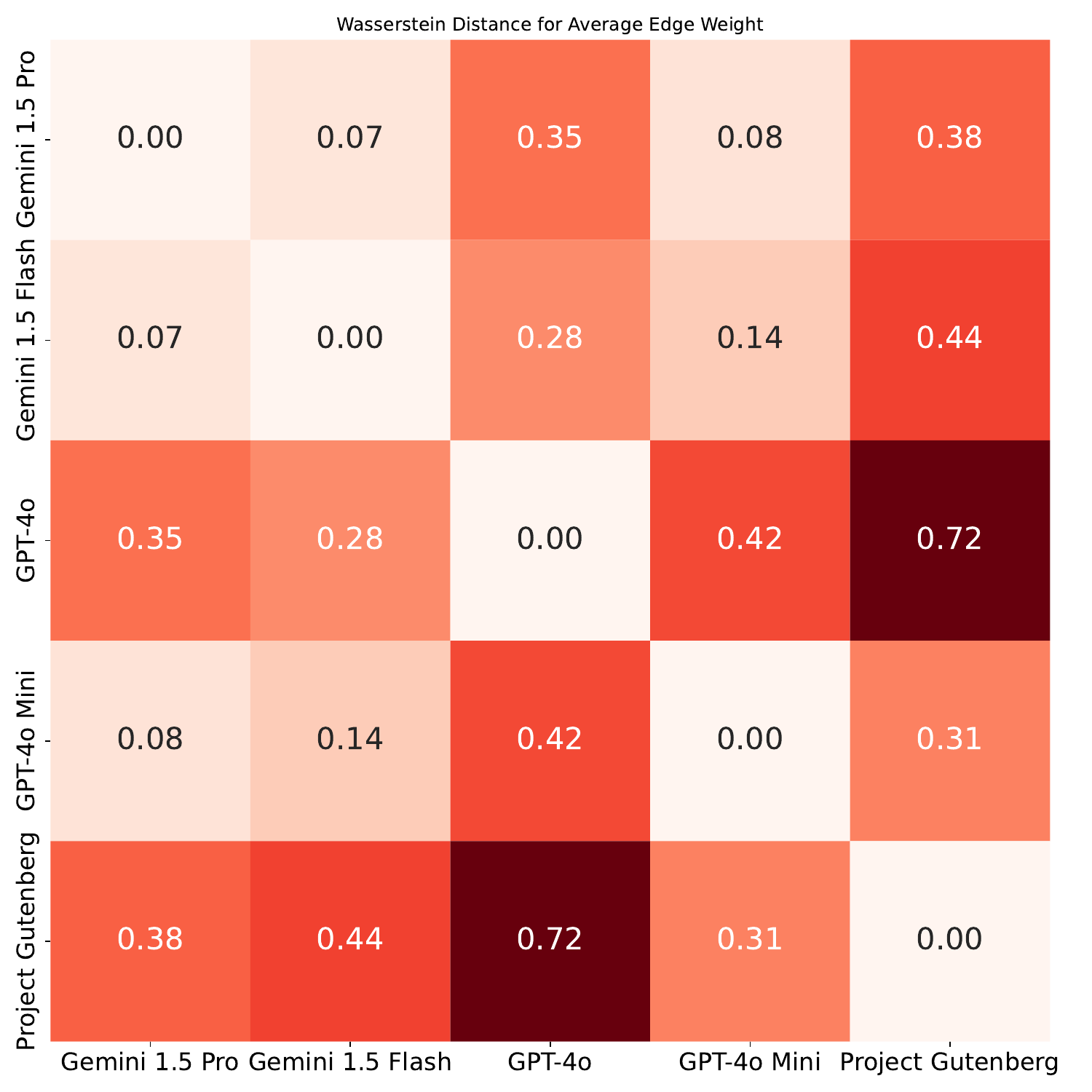}
    \caption{WD for Average Edge Weight.}
\end{subfigure}
\begin{subfigure}{0.49\textwidth}
    \includegraphics[width=\linewidth]{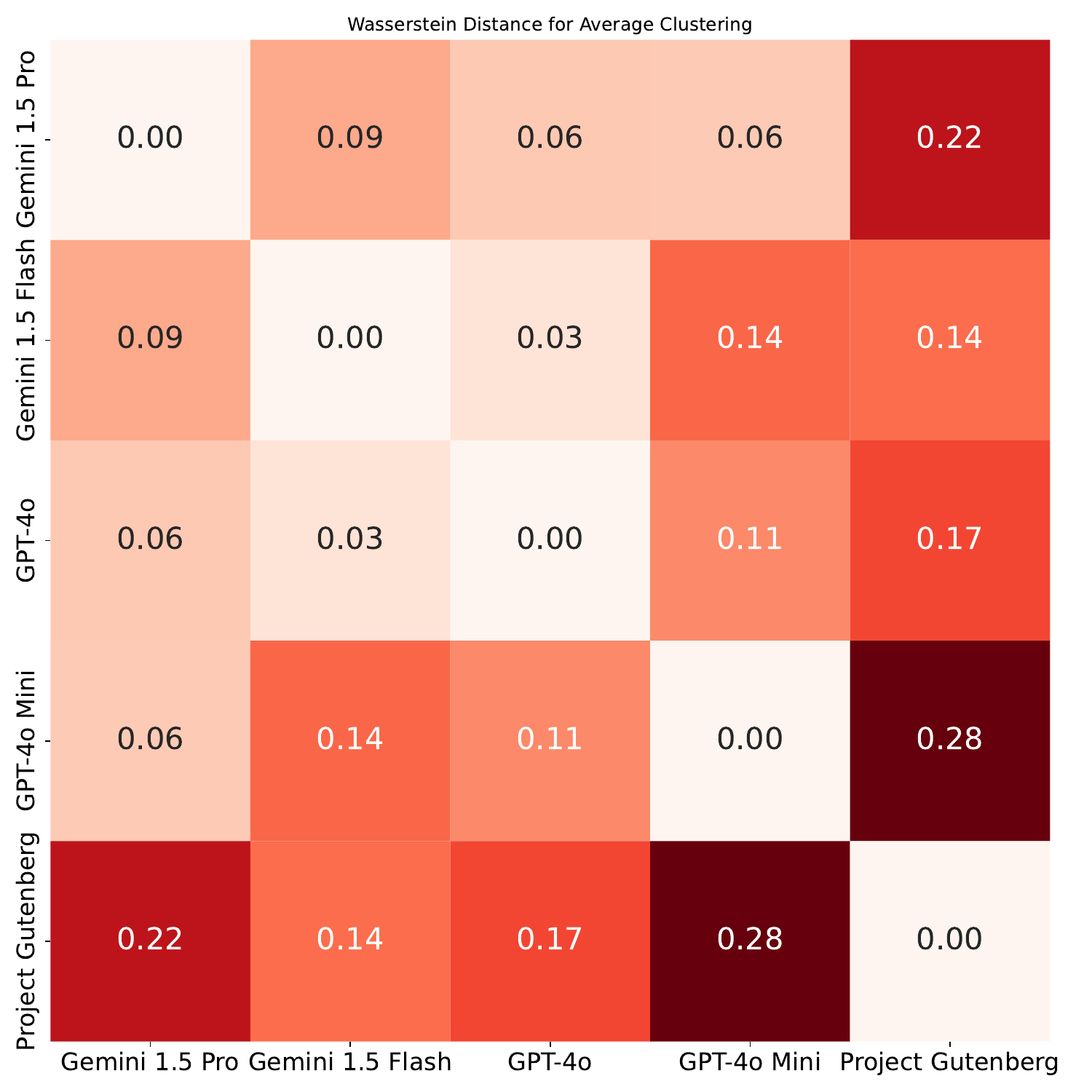}
    \caption{WD for Average Clustering Coefficient.}
\end{subfigure}
\begin{subfigure}{0.49\textwidth}
    \includegraphics[width=\linewidth]{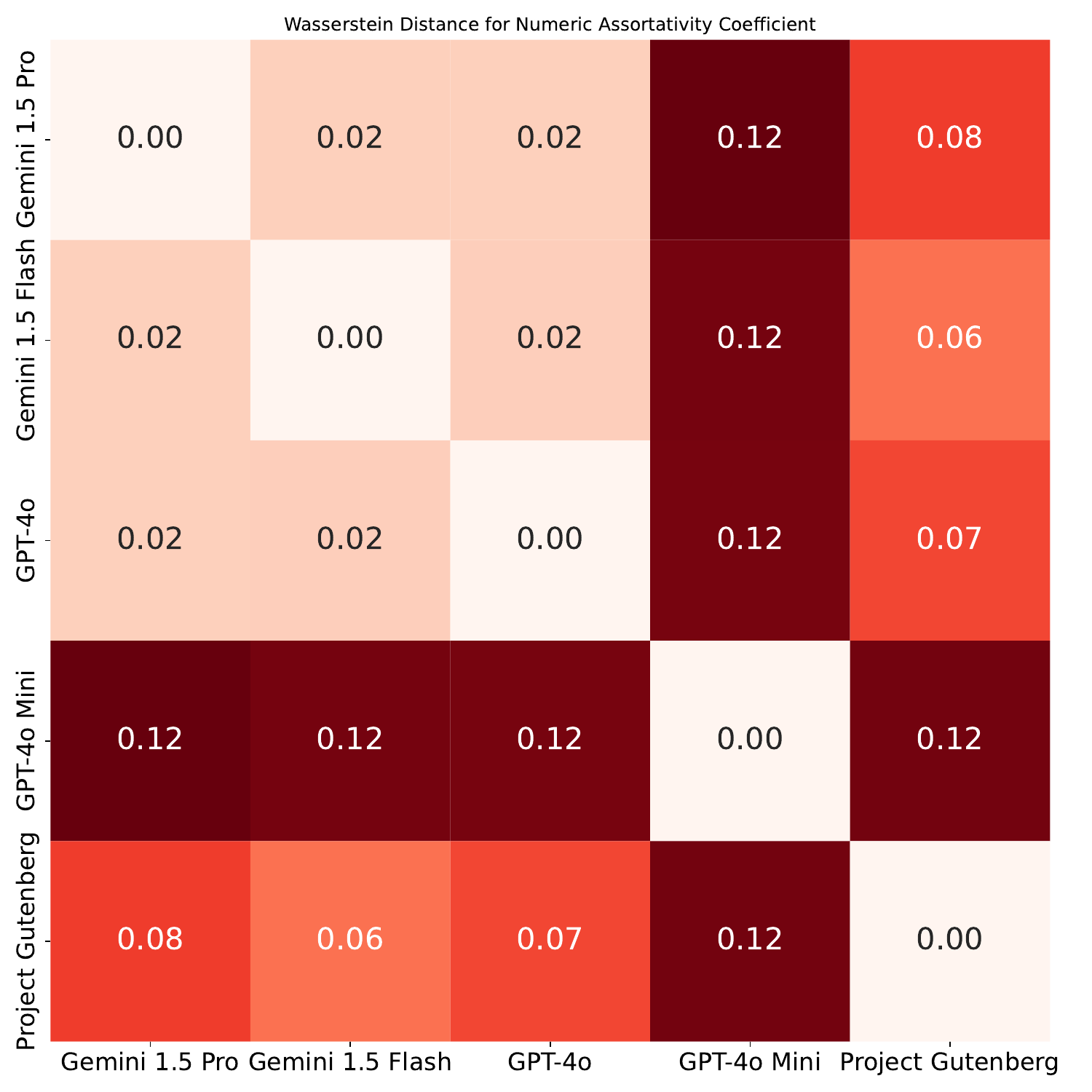}
    \caption{WD for Assortativity Mixing.}
\end{subfigure}
\end{figure}

\clearpage

\section{t-test} \label{appendix:t-test}
We ran Welch's t-tests for two independently-sampled sets of scores, assuming that the variances of the two sets of samples differ, with \texttt{scipy.stats.ttest\_ind} function. The null hypothesis is $H_0: \mu_{m_{LLM1}} = \mu_{m_{LLM2}}$. High p-values indicate that, at a certain statistical significance level, we \textit{cannot} reject the null hypothesis that the means of the two score sets from different models are identical. For every metric except for the assortativity mixing and the average clustering of negative networks, the sample size of scores for each writer was equal to the number of networks analyzed (GPT 4o: $n=251$, GPT 4o Mini: $n=249$, Gemini 1.5 Pro: $n=252$, Gemini 1.5 Flash: $n=249$, and Humans: $n=168$). We have smaller sample sizes for the two metrics above due to the system's inability to calculate them for some networks. In assortativity mixing, the sample size of humans is $n=167$, and the other sample sizes are equal to their network counts. For the clustering coefficient of negative networks, GPT 4o has $n=245$, Gemini 1.5 Flash has $n=247$, humans have $n=166$, and GPT 4o Mini, Gemini 1.5 Pro do not have any missing instances.

Several metrics across some models, such as density (Gemini Pro and GPT 4o: $p=0.520$) and average clustering of positive networks (Gemini Pro and GPT 4o Mini: $p=0.792$, GPT 4o and GPT 4o Mini: $p=0.116$) and negative networks (Gemini Flash and Pro: $p=0.840$), have high p-values, indicating that the score samples from two distinct models are not unlikely to be drawn from the same sample space. Interestingly, only the assortativity mixing consistently shows high p-values with a couple of pairs that include humans (Gemini Flash and GPT 4o: $p=0.852$, Gemini Flash and Pro: $p=0.607$, Gemini Pro and GPT 4o: $p=0.736$, Gemini Flash and Humans: $p=0.165$, GPT 4o and Humans: $p=0.122$). It is also noteworthy that, overall, the density of negative networks has high p-values compared to positive networks (GPT 4o and GPT 4o Mini: $p=0.936$, Gemini Flash and GPT 4o: $p=0.546$, Gemini Flash and GPT 4o Mini: $p=0.369$). Besides assortativity mixing, as expected, p-values for pairs including human-written stories are consistently very low ($p<0.01$), except for the density of positive networks with Gemini 1.5 Flash ($p = 0.070$), which still indicates the weak evidence for the null hypothesis.

\end{document}